\documentclass[9pt,conference]{IEEEtran}
\IEEEoverridecommandlockouts
\pdfoutput=1

\usepackage{amsmath}
\usepackage{dsfont}
\usepackage{graphicx, graphics, theorem, times, amsfonts, amsmath, amssymb, cite}
\usepackage{tikz}
\usetikzlibrary{shapes,arrows}
\usepackage{pgfplots}
\usepackage{xcolor}
\usepackage{hyperref}
\usepackage{multirow}
\usepackage{rotating}
\usepackage{cleveref, subcaption}
\usepackage{bm}
\input{mysymbol.sty}
\usepackage[utf8]{inputenc}
\usepackage{enumitem}
\usepackage{algorithm,algpseudocode}
\usepackage{array}
\usepackage{moresize} 

\newcolumntype{P}[1]{>{\centering\arraybackslash}p{#1}}

\newcommand{\QED}{\hfill\ensuremath{\blacksquare}}

\newcommand{\alna}[1]{\begin{alignat}{3}&#1&\end{alignat}}

\newtheorem{mytheorem}{\hspace{-1pt}\bf Theorem}

\def \bbCo {\bbC_{{\scriptscriptstyle \ccalO}}}

\def \bbCh {\bbC_{{\scriptscriptstyle \ccalH}}}
\def \bbCoh {\bbC_{{\scriptscriptstyle \ccalO\ccalH}}}
\def \bbCho {\bbC_{{\scriptscriptstyle \ccalH\ccalO}}}

\def \hbCo {\hbC_{{\scriptscriptstyle \ccalO}}}
\def \hbCoz {\hbC_{{\scriptscriptstyle \ccalO, 0}}}
\def \hbCot {\hbC_{{\scriptscriptstyle \ccalO, t}}}
\def \hbCoto {\hbC_{{\scriptscriptstyle \ccalO, {t-1}}}}

\def \bbXo {\bbX_{{\scriptscriptstyle \ccalO}}}
\def \bbXh {\bbX_{{\scriptscriptstyle \ccalH}}}

\def \bbxot {\bbx_{{\scriptscriptstyle \ccalO,t}}}

\def \bbxoi {\bbx_{{\scriptscriptstyle \ccalO, i }}}

\def \bbBt {\bbB_{{\scriptscriptstyle t}}}
\def \bbDt {\bbB_{{\scriptscriptstyle t}}}

\def \bbSo {\bbS_{{\scriptscriptstyle \ccalO}}}
\def \bbSot {\bbS_{{\scriptscriptstyle \ccalO, t}}}

\def \bbSotn {\bbS_{{\scriptscriptstyle \ccalO, t+1 }}}
\def \bbSoz {\bbS_{{\scriptscriptstyle \ccalO, 0}}}
\def \bbSh {\bbS_{{\scriptscriptstyle \ccalH}}}
\def \bbSoh {\bbS_{{\scriptscriptstyle \ccalO\ccalH}}}
\def \bbSho {\bbS_{{\scriptscriptstyle \ccalH\ccalO}}}

\def \hbSo {\hbS_{{\scriptscriptstyle \ccalO}}}
\def \hbSot {\hbS_{{\scriptscriptstyle \ccalO, t}}}
\def \hbSoto {\hbS_{{\scriptscriptstyle \ccalO, t-1}}}
\def \hbSotz {\hbS_{{\scriptscriptstyle \ccalO, 0}}}
\def \hbSotn {\hbS_{{\scriptscriptstyle \ccalO, t+1 }}}
\def \hbSoz {\hbS_{{\scriptscriptstyle \ccalO, 0}}}

\def \bbPt {\bbP_{{\scriptscriptstyle t}}}

\def \bbPtn {\bbP_{{\scriptscriptstyle t+1 }}}
\def \bbPz {\bbP_{{\scriptscriptstyle 0}}}

\def \hbPt {\hbP_{{\scriptscriptstyle t}}}

\def \hbPtn {\hbP_{{\scriptscriptstyle t+1 }}}
\def \hbPz {\hbP_{{\scriptscriptstyle 0}}}

\newcommand{\norm}[1]{\left\lVert#1\right\rVert}

     \makeatother

     \makeatletter
     \renewcommand\small{%
     \@setfontsize\small\@ixpt{11}%
     \abovedisplayskip 3.5\p@ \@plus3\p@ \@minus2\p@
     \abovedisplayshortskip \z@ \@plus2\p@
     \belowdisplayshortskip 3\p@ \@plus2\p@ \@minus2\p@
     \def\@listi{\leftmargin\leftmargini
     \topsep 4\p@ \@plus2\p@ \@minus2\p@
     \parsep 2\p@ \@plus\p@
     \@minus\p@
     \itemsep
     \parsep}%
     \belowdisplayskip
     \abovedisplayskip
     }
     \makeatother

\pgfplotsset{compat=1.17}
\pgfplotstableset{col sep=comma}
\usepgfplotslibrary{fillbetween}
\tikzset{every mark/.append style={scale=1.5, solid}, font=\footnotesize}
\pgfplotsset{
    width=1.05\textwidth,
    legend style={
        font=\ssmall ,  
        inner xsep=1pt,
        inner ysep=1pt,
        nodes={inner sep=1pt}},
    legend cell align=left,
	every axis/.append style={line width=0.5pt},
	every axis plot/.append style={line width=1.5pt},
    every axis y label/.append style={yshift=-5pt}
}

\begin{document}

\title{Online Network Inference from Graph-Stationary Signals \\ with Hidden Nodes
\thanks{This work was partially supported by the Spanish AEI PID2022-136887NB-I00, the Community of Madrid via the ELLIS Madrid Unit, and the U.S. NSF under award CCF-2340481. 
Research was sponsored by the Army Research Office and was accomplished under Grant Number W911NF-17-S-0002. The views and conclusions contained in this document are those of the authors and should  not be interpreted as representing the official policies, either expressed or implied, of the Army Research Office or the U.S. Army or the U.S. Government. The U.S. Government is authorized to reproduce and distribute reprints for Government purposes notwithstanding any copyright notation herein.
        \href{mailto:andrei.buciulea@urjc.es}{andrei.buciulea@urjc.es}, 
        \href{mailto:nav@rice.edu}{nav@rice.edu},
        \href{mailto:samuel.rey.escudero@urjc.es}{samuel.rey.escudero@urjc.es},
        \href{mailto:segarra@rice.edu}{segarra@rice.edu},
        \href{mailto:antonio.garcia.marques@urjc.es}{antonio.garcia.marques@urjc.es}
        }
}

\author{
\IEEEauthorblockN{
Andrei Buciulea\IEEEauthorrefmark{1},
Madeline Navarro\IEEEauthorrefmark{2},
Samuel Rey\IEEEauthorrefmark{1},
Santiago Segarra\IEEEauthorrefmark{2},
Antonio G. Marques\IEEEauthorrefmark{1},
} %
\IEEEauthorblockA{
\IEEEauthorrefmark{1}Dept. of Signal Theory and Communications, Rey Juan Carlos University, Madrid, Spain } %
\IEEEauthorblockA{
\IEEEauthorrefmark{2}Dept. of Electrical and Computer Engineering, Rice University, Houston, TX, USA } %
}

\maketitle

\begin{abstract}
    Graph learning is the fundamental task of estimating unknown graph connectivity from available data.
    Typical approaches assume that not only is all information available simultaneously but also that all nodes can be observed.
    However, in many real-world scenarios, data can neither be known completely nor obtained all at once.
    We present a novel method for online graph estimation that accounts for the presence of hidden nodes.
    We consider signals that are stationary on the underlying graph, which provides a model for the unknown connections to hidden nodes.
    We then formulate a convex optimization problem for graph learning from streaming, incomplete graph signals.
    We solve the proposed problem through an efficient proximal gradient algorithm that can run in real-time as data arrives sequentially.
    Additionally, we provide theoretical conditions under which our online algorithm is similar to batch-wise solutions.
    Through experimental results on synthetic and real-world data, we demonstrate the viability of our approach for online graph learning in the presence of missing observations.
\end{abstract}

\begin{IEEEkeywords}
    Online graph learning, graph signal processing, hidden nodes.
\end{IEEEkeywords}

\section{Introduction}\label{sec:introduction}










Estimating the unknown network topology from data for modeling complex systems is a crucial step for several signal processing and machine learning tasks~\cite{mateos2019connecting,dong_2019_learning}.
Methods to estimate the graph structure assume a model describing pair-wise node-to-node interactions that explain the behavior of data on the graph.
The graph topology can then be inferred from observed data through algebraic and statistical methods.
Classical examples include correlation-based methods~\cite[Ch. 7.3.1]{kolaczyk2009book}, graphical lasso (GL)~\cite{GLasso2008,danaher2014joint,buciulea2024polynomial}, and GSP-based models, which exploit signal properties such as smoothness or graph stationarity ~\cite{Kalofolias2016inference_smoothAISTATS16,egilmez2017graph,segarra2017network,buciulea2022learning}.
Other works consider more varied and potentially more realistic scenarios, such as graph learning in multilayer settings~\cite{navarro2022JointInferenceMultiple,rey2022joint}, from incomplete or streaming graph signals~\cite{navarro2024joint,zhang2025graphlearningincomplete,shafipour2020online}, or for more complex structural priors~\cite{navarro2022joint,rey2023enhanced}.

Realistically, graph data cannot always be collected as batches of fully observed data.
The occlusion of a subset of nodes may cause critical information loss for graph learning, even for connections only between observed nodes~\cite{buciulea2022learning}.
Moreover, there is an increasing need to learn graphs from data that arrive in real-time, such as for communication networks, evolving social networks, and brain networks observed over time~\cite{marano2016importance,liegeois2020revisiting}.
Indeed, even beyond these common applications, there is recent demand for large amounts of data that are difficult to collect and parse all at once.
Many works infer potentially time-varying graphs from dynamic signals, but estimation is typically performed offline on a batch of signals~\cite{zhang2025graphlearningincomplete}.
Online network inference has been proposed for different graph signal models, including classical GSP models such as smooth signals, stationary signals~\cite{djuric2018cooperative,marques2017stationary}, and Gaussian Markov random fields, see, e.g., ~\cite{saboksayr2023dualbased,natali2022learning,shafipour2020online,ye2024timevaryinggraphlearning} for recent online approaches.
However, while existing works consider network inference from streaming or incomplete data, the consideration of both limitations (i.e., streaming \emph{and} incomplete data) has not been addressed under graph signal stationarity~\cite{zaman2023onlinejoint, money2022online, cirillo2023estimating}.

In this work, we propose online graph learning from stationary graph signals in the presence of hidden nodes.
By modeling the relationship between observed and hidden nodes under graph stationarity, we formulate a convex optimization problem for estimating the connectivity among observed nodes while accounting for relevant information from hidden nodes.
We present an algorithm based on proximal gradient descent to solve the problem given either a batch or a stream of nodal observations~\cite{shafipour2020online}.
While our emphasis is on the more challenging task of online network inference, the proposed algorithm enjoys greater efficiency than existing approaches that learn graphs from stationary graph signals that are incomplete~\cite{buciulea2022learning}.
Our contributions are as follows.

\begin{itemize}[left= 5pt .. 15pt, noitemsep]
\item[1)] 
We present a convex optimization problem for online graph learning from stationary graph signals for which a subset of nodes are hidden.
\item[2)]
We provide a proximal gradient algorithm for solving the proposed problem, which not only accomplishes online graph learning but also improves upon existing methods for estimation from incomplete stationary graph signals.
\item[3)]
We further provide theoretical guarantees for the performance of our method for estimating both the time-varying subgraph of observed nodes and accounting for the connections between observed and hidden nodes.
In particular, we show that our online approach can track a time-varying solution obtained by batch-wise estimation.
\item[4)]
Through empirical simulations, we demonstrate that our approach is valid for online network inference not only when graph signals are obtained sequentially but also when the underlying graph varies over time.
\end{itemize}


\section{Background}\label{S:netw_reconstruction}

\subsection{Graph signal processing}\label{Ss:gsp}

We are interested in recovering the structure of an undirected graph $\ccalG = (\ccalV,\ccalE)$, where $\ccalV = \{1,\dots, N\}$ is a set of $N$ nodes and $\ccalE \subseteq \{  (i,j) | i,j\in\ccalV\}$ is the set of edges connecting pairs of nodes.
Moreover, consider $R$ observations $\bbx_i \in \reals^N$ for $i=1,\dots,R$, where each entry of $\bbx_i$ corresponds to a node in the graph.
Estimating the graph topology from the data $\bbX = [\bbx_1,\dots,\bbx_R]\in\reals^{N\times R}$ requires leveraging a model for the relationship between the data and the graph $\ccalG$. As we are primarily interested in the connectivity patterns of $\ccalG$, we estimate an algebraic representation of the graph known as its graph shift operator (GSO) $\bbS \in \reals^{N \times N}$~\cite{shuman2013emerging,sandryhaila2013discrete,djuric2018cooperative}, a matrix whose sparsity patterns inform the connectivity pattern of $\ccalG$. In particular, $S_{ij} = S_{ji}\neq 0$ if and only if $i=j$ or $(i,j)\in\ccalE$.
Then the topology of the graph can be estimated by solving an inverse problem of the form $\bbS = f^{-1}(\bbX)$, where $f$ denotes the model for the signal realizations $\bbX$ as a function of $\bbS$.
Additional assumptions can be added to the problem by considering symmetric $\bbS$  as we consider undirected graphs or different choices of GSO including the adjacency matrix or the graph Laplacian~\cite{djuric2018cooperative}.

We consider graph signals as stochastic realizations of a process $\bbx$ that is stationary on $\ccalG$~\cite{djuric2018cooperative,marques2017stationary}.
Graph stationarity encompasses several existing models such as correlation networks and Markov random fields~\cite{segarra2017network,pasdeloup2018characterization}, and many applications can be naturally modeled using stationary graph signals~\cite{thanou2017learning,li2019identifying}.
One consequence is that the \textit{covariance matrix} of the graph signals $\bbC = \mbE[\bbx\bbx^\top]$ can be written as a \textit{polynomial of} the GSO $\bbS$~\cite{marques2017stationary}.
This yields the critical fact that the GSO and the covariance commute, that is, $\bbC\bbS = \bbS\bbC$, a relationship commonly employed for network inference~\cite{segarra2017network,shafipour2020online,buciulea2022learning}.
Moreover, the commutativity between $\bbC$ and $\bbS$ allows us to conveniently model the influence of hidden nodes.

\subsection{Incomplete and streaming graph data}\label{Ss:hid}

Recall that we assume the graph signals are partially observed, meaning we do not have access to the entire data matrix $\bbX$. Instead, data is measured on a subset of nodes $\ccalO \subset \ccalV$ consisting of $O$ nodes, while the remaining subset $\ccalH = \ccalV \backslash \ccalO$, containing $H$ nodes, is unobserved.
Specifically, we observe the data $\bbXo \in \reals^{O \times R}$, while the data $\bbXh \in \reals^{H \times R}$ remains hidden.
We can then partition $\bbS$, $\bbC$, and $\bbX$ conformally based on the pairwise relationships between the data from observed and hidden nodes as follows
\alna{
    \bbS = \begin{bmatrix}
        \bbSo & \bbSoh \\ \bbSho & \bbSh
    \end{bmatrix}, 
    \quad 
    \bbC = \begin{bmatrix}
        \bbCo & \bbCoh \\ \bbCho & \bbCh
    \end{bmatrix}, 
    \quad 
    \bbX = \begin{bmatrix}
        \bbXo \\ \bbXh
    \end{bmatrix},
\label{e:blocks}
}
where $\bbSoh = \bbSho^\top$ and $\bbCoh=\bbCho^\top$ since $\bbS$ is symmetric. 
The submatrices $\bbSo$, $\bbCo \in \reals^{O \times O}$ consider pair-wise relationships between observed nodes and $\bbSh$, $\bbCh \in \reals^{H \times H}$ between hidden nodes, while $\bbSoh, \bbCoh \in \reals^{O \times H}$ model relationships between pairs of observed and hidden nodes.
In this setting, the goal is to estimate the connectivity among the observed nodes $\ccalO$, that is, to estimate the submatrix $\bbSo$.
To ensure that the problem is feasible, we require that $H \ll O$, where there are far more observed nodes than hidden.
Due to the convenience of the blockwise structure in~\eqref{e:blocks}, previous works have considered this problem in different scenarios
such as for estimating multiple graphs~\cite{navarro2022JointInferenceMultiple}, for time-varying graph data~\cite{ye2024timevaryinggraphlearning}, or for additional assumptions on the graph signals~\cite{zhang2025graphlearningincomplete}.

However, existing methods to estimate $\bbSo$ do so in a batch-wise and offline manner, where all data in $\bbXo$ are available simultaneously.
In an online setting, accurate estimation is more difficult, which can exacerbate discrepancies due to missing nodal information.
Previous works consider estimating dynamic graphs from partial observations, which primarily consider different signal models than ours and thus require different approaches~\cite{zaman2023onlinejoint,ioannidis2019semiblind,money2022online,zhang2025graphlearningincomplete}.
To emphasize the value of considering stationarity, we point to the work in~\cite{shafipour2020online}, which provides an efficient online algorithm for estimating a graph from streaming stationary graph signals.
Thus, we extend this to the incomplete data setting, for which we exploit our signal model with a convenient formulation for hidden information~\cite{buciulea2022learning}.
We then obtain interpretable theoretical guarantees for tracking the optimal time-varying solution with fewer required assumptions than existing works~\cite{shafipour2020online,zhang2025graphlearningincomplete}.

\section{Graph learning with hidden nodes}\label{S:method}

First, we recount the graph learning problem from stationary graph signals in the presence of hidden nodes~\cite{buciulea2022learning}.
Recall that graph stationarity implies that the matrices $\bbC\bbS$ and $\bbS\bbC$ must be equal. Focusing on the upper $O \times O$ block of these two matrices and leveraging the partitions in~\eqref{e:blocks}, it readily follows that $\bbCo\bbSo + \bbCoh\bbSho = \bbSo\bbCo +\bbSoh\bbCho$. Upon defining $\bbP := \bbCoh\bbSho$ and recalling that $\bbC$ and $\bbS$ are symmetric, we have that stationarity implies
\begin{equation}\label{e:commut}
    \bbCo\bbSo + \bbP  = \bbSo\bbCo + \bbP^\top.
\end{equation}
Thus, estimating the structure in $\bbSo$ from incomplete stationary graph signals can be posed as the joint inference of $\bbSo$ and $\bbP$.
In particular, we estimate the observed covariance submatrix as $\hbCo = \frac{1}{R} \bbXo\bbXo^\top$ and solve
\alna{
    \min_{\bbSo, \bbP} && ~~~
    \norm{\bbSo}_1
    +
    \rho \norm{\bbP}_{2,1}
&\nonumber\\&
    ~ \mathrm{s.t.} && ~~~
    \| \hbCo \bbSo + \bbP - \bbSo \hbCo - \bbP^\top \|_F \leq \epsilon, 
    \quad 
    \bbSo \in \ccalS, ~
\label{e:opt_batch}
}
where the first constraint approximates the relationship in~\eqref{e:commut}, while the second imposes valid GSOs according to the set $\ccalS$.
We further promote parsimonious estimates $\bbSo$ for ease of interpretation and downstream computation, thus the first term in the objective of~\eqref{e:opt_batch} encourages a sparse estimate.
Moreover, as $\bbP$ aims to estimate $\bbCoh \bbSho$, then $\bbP$ is not only low rank as $H \ll O$, but the sparsity of $\bbSho$ ought to yield a column sparse $\bbP$.
Thus, the $\ell_{2,1}$ norm penalty in~\eqref{e:opt_batch} similarly encourages parsimonious estimation through the matrix $\bbP$.
The tradeoff between the two terms is controlled by $\rho > 0$.

\subsection{Online graph learning}\label{Ss:online}

While the formulation in~\eqref{e:opt_batch} is successful for batch-wise estimation, we are interested in an online setting, where data arrives sequentially. 
In particular, at each time instant $t \in \naturals$, we obtain a new partially observed graph signal $\bbxot$ on the nodes $\ccalO$.
We then aim to solve a time-varying problem,
\alna{
    \bbSot^*, \bbPt^* ~~=~~ &\argmin_{\bbSo,\bbP} & ~~~
    \|\bbSo\|_1 + \rho \|\bbP\|_{2,1} + g_t( \bbSo, \bbP )
    &\nonumber\\&
    & \mathrm{s.t.}~~ & ~~~
    \bbSo = \bbSo^\top, ~~
    \mathrm{diag}(\bbSo) = \bbzero, ~~
    \bbSo \geq 0, 
&\nonumber\\&
    && ~~~
    [\bbSo]_{\Omega} = \bbS_{\Omega},
\label{e:opt_online}
}
where 
\begin{equation}\label{e:commut_penalty}
    g_t(\bbSo,\bbP) = \frac{\mu}{2} \| \hbCot \bbSo + \bbP - \bbSo\hbCot - \bbP^\top \|_F^2,
\end{equation}
that is, we replace the commutativity constraint with a penalty in the objective function.
Indeed, based on duality theory, for every $\epsilon > 0$ there exists a value of $\mu > 0$ such that the penalty $g_t$ yields an equivalent solution to that of the constraint in~\eqref{e:opt_batch}. 
Moreover, as data is observed sequentially, the notation $\hbCot$ denotes the estimate at time $t$ of the covariance submatrix based on all available signals $\{\bbxoi\}_{i=1}^t$. The specific expression for $\hbCot$ will be introduced in the next subsection, where we propose an efficient approach to solve the online problem while considering hidden nodes.
Finally, observe that we specify $\ccalS$ to estimate $\bbSo$ as unweighted, unsigned adjacency submatrices.
The problem~\eqref{e:opt_online} is suitable for other GSOs, but we exemplify our approach with adjacency matrices to specify the update steps for our ensuing algorithm. 
For the final constraint, we assume prior knowledge for a subset of edges $\Omega\subset \ccalV\times\ccalV$.
Indeed, as we consider a time-varying setting, it is reasonable to assume some historical information, such as prior knowledge of relationships in an evolving social network.
Moreover, assuming a nonempty $\Omega$ allows us to preclude the trivial solution $\bbSo = \bbzero_{O\times O}$ without requiring constraints that may lead to infeasible solutions~\cite{shafipour2020online}.
We let $\bbS_{\Omega}$ contain the values of the known entries of $\bbSo$.

A naive approach for sequentially arriving graph signals is to estimate $\bbSot^*$, $\bbPt^*$ in an offline manner, solving~\eqref{e:opt_online} at every time $t$ using all available data $\{\bbxoi\}_{i=1}^t$.
However, this is inefficient as data will likely arrive more quickly than~\eqref{e:opt_online} can be solved.
Moreover, in settings where changes in the graph take place with high frequency, it might not even be prudent to solve \eqref{e:opt_online} with high precision.
Indeed, this poses a trade-off where sub-optimal and fast solutions may be preferred over optimal but time-consuming alternatives.

Instead, we consider an efficient approach based on an online proximal algorithm~\cite{shafipour2020online}, where we perform one iteration of proximal gradient descent at each time $t$.
Not only can we then perform online optimization, but we also enjoy the added benefit of more efficiently tracking a time-varying graph.
Indeed, a significant distribution shift upon the arrival of $\bbxot$ may yield an unstable estimate $\bbSot^*$ if obtained in a batch-wise manner, while online inference can encourage smoother changes from the previous estimate $\hbSoto$.

\subsection{Proximal gradient algorithm}\label{Ss:pgd}

We next derive the computations necessary for our proposed algorithm.
First, at each time $t$ a new signal $\bbxot$ arrives, and we apply a recursive update to the sample covariance submatrix $\hbCot$ through a rank-one correction as follows
\begin{equation}\label{e:cov_upd}
    \hbCot = \frac{1}{t} \left(
        (t-1)\hbCoto + \bbxot\bbxot^\top
    \right).
\end{equation}
Then, we perform a gradient descent step for $\bbSot$ and $\bbPt$.
We thus compute the gradient of $g_t$ with respect to $\bbSo$ as
\alna{
    \nabla_{\bbSo} g_t (\bbSo,\bbP)
    &~=~&
    \mu [ \hbCot (\hbCot \bbSo + \bbP - \bbSo \hbCot - \bbP^\top )
&\nonumber\\&
    && 
    -
    (\hbCot \bbSo + \bbP - \bbSo \hbCot - \bbP^\top ) \hbCot ],
\label{e:grad_g_S}
}
where we recall that $\mu$ is the weight of the conmutativity regularizer. The gradient with respect to  $\bbP$ is
\alna{
    \nabla_{\bbP} g_t (\bbSo,\bbP)
    =
    2\mu (\hbCot \bbSo + \bbP - \bbSo \hbCot - \bbP^\top ).
\label{e:grad_g_P}
}
Following gradient descent, we require proximal operators for $\bbSo$ and $\bbP$, which account for the remaining terms in the objective and project the estimate $\bbSo$ onto the feasible set $\ccalS$~\cite{shafipour2020online}.
Each entry of the corresponding proximal operator for $\bbSo$ follows
\alna{
    [ \mathrm{prox}_{ \gamma \| \cdot \|_1, \ccalS } ( \bbQ ) ]_{ij}
    =
    \begin{Bmatrix}
        0, & i=j \\
        [\bbS_{\Omega}]_{ij}, & (i,j)\in\Omega \\
        \max( 0, Q_{ij} - \gamma ), & \mathrm{otherwise}
    \end{Bmatrix}
\label{e:prox_S}
}
for every $i,j=1,\dots, O$ and some $\gamma>0$.
The proximal operator for $\bbP$ can be computed for each column $j=1,\dots,O$ as
\alna{
    [\mathrm{prox}_{\gamma \|\cdot\|_{2,1}} (\bbQ)]_{\cdot,j}
    =
    \max\left( 0, 1 - \frac{\gamma}{ \|\bbQ_{\cdot,j}\|_2 } \right) \bbQ_{\cdot,j}
\label{e:prox_P}
}
for some $\gamma > 0$.
With the above derivations, the proposed method is summarized in Alg.~\ref{alg:PG}.
The online proximal gradient steps require choice of a time-varying step size $\gamma_t$ for every $t\in\naturals.$
Given that the gradient of the objective function in \eqref{e:opt_online} is Lipschitz continuous~\cite{shafipour2020online}, a reasonable choice for $\gamma_t$ is one that satisfies $\gamma_t < (2 \mu \sigma_t^2)^{-1}$, where $\sigma_t$ is the largest singular value of $\hbCot$. 

We elaborate on some advantages of our algorithm.
First, observe that the feasible set $\ccalS$ is closed under~\eqref{e:prox_S}, so as long as the initialization $\hbSotz$ is feasible, then subsequent iterations will yield feasible estimates $\hbSot$ for every $t\in\naturals$.
Second, we allow for prior data, that is, we may have an initial estimate of the covariance submatrix $\hbCoz$, which may come from an initially observed batch of signals.
Third, note that we can apply the proximal gradient steps 4 to 10 of Alg.~\ref{alg:PG} given a full batch of samples, which yields an efficient algorithm for solving~\eqref{e:opt_batch}.
Indeed, notwithstanding the effectiveness of Alg.~\ref{alg:PG} for online graph learning, we also provide an optimization approach for solving~\eqref{e:opt_batch}, while previous works relied on off-the-shelf solvers for problems of this form~\cite{buciulea2022learning,navarro2022JointInferenceMultiple}.
Finally, by the convex formulation of~\eqref{e:opt_online}, we theoretically show that the online solution from Alg.~\ref{alg:PG} can track the ideal batch-wise solution to~\eqref{e:opt_online} over time $t$, analogous to the result in~\cite{shafipour2020online}.

\vspace{-.1cm}

\begin{mytheorem}\label{thm:conv}
    Let $v_t := \norm{ \bbSotn^* - \bbSot^* }_F + \norm{ \bbPtn^* - \bbPt^* }_F$ denote the variability of the offline solution in~\eqref{e:opt_online}.
    Then, for all $t\in\naturals$, the iterates $\hbSot$, $\hbPt$ obtained from Alg.~\ref{alg:PG} satisfy
    \alna{
        \| \hbSot-\bbSot^* \|_F +
        \| \hbPt-\bbPt^* \|_F
    &\nonumber\\&
        \quad 
        \leq
        L_{t-1} \left(
            \|\hbSoz - \bbSoz^* \|_F +
            \|\hbPz - \bbPz^* \|_F +
            \sum_{i=0}^{t-1} \frac{v_i}{L_i}
        \right),
    \label{e:thm1_bnd}
    }
    for small enough $\mu>0$, where $L_t = \prod_{i=0}^t 5\gamma_i (\sigma_i+1)^2$.
\end{mytheorem}

Due to lack of space, we provide a brief proof sketch.
We bound the differences $\|\hbSotn-\bbSot^*\|_F$ and $\| \hbPtn - \bbPt^* \|_F$ by the non-expansiveness of the proximal operators and the Lipschitz smoothness of $\nabla_{\bbSo} g_t$ and $\nabla_{\bbP}g_t$.
Then, by the triangle inequality and the definition of $v_t$, we obtain the right-hand side of~\eqref{e:thm1_bnd} by recursion.
Theorem~\ref{thm:conv} shows our algorithm can approximate the inefficient but ideal batch-wise solution in~\eqref{e:opt_online}.
In particular, we can ensure that $\hbSot$ and $\hbPt$ track the solutions $\bbSot^*$ and $\bbPt^*$ as long as they do not vary too rapidly over time.
Thus, even under the setting of time-varying graphs, sequentially arriving data, and missing nodes, our algorithm enjoys efficient optimization with provable tracking of the ideal batch-wise solution.
As a remark, we highlight that performing multiple iterations of steps 4 to 9 for each time step $t$ can enhance the accuracy of graph estimation by iteratively refining the solution, though at the cost of increased computational complexity. This trade-off will be examined in greater detail in the subsequent section.


\begin{algorithm}[t]
\caption{Online graph learning via proximal gradient descent}
\label{alg:PG}

\begin{algorithmic}[1]
\Require $\hbCoz$, $\mu > 0$, $\rho > 0$
\State Initialize $\hbSoz$, $\hbPz$, and $\gamma_0 < (2\mu\sigma_0^2)^{-1}$
\For{$t \in \naturals$}
    \State Update $\hbCot$ via~\eqref{e:cov_upd} and $\gamma_t < (2\mu\sigma_t^2)^{-1}$
    \State Compute $\nabla_{\bbSo} g_t (\hbSot)$ via~\eqref{e:grad_g_S}
    \State Take gradient descent step $\bbBt = \hbSot - \gamma_t \nabla_{\bbSo} g_t(\hbSot)$
    \State Update $\hbSotn = \mathrm{prox}_{\gamma_t \|\cdot\|_1,\ccalS} \left( \bbBt\right)  $ via \eqref{e:prox_S}
    \State Compute $\nabla_{\bbP} g(\hbPt)$ via \eqref{e:grad_g_P}
    \State Take gradient descent step $\bbDt = \hbPt - \gamma_t \nabla_{\bbP} g_t(\hbPt)$
    \State Update $\hbPtn = \mathrm{prox}_{\gamma_t\rho \|\cdot\|_{2,1}} \left( \bbDt\right)$ via~\eqref{e:prox_P}
    \State $t = t + 1$
\EndFor
\State \Return $\hbSotn$, $\hbPtn$
\end{algorithmic}

\end{algorithm}


\section{Numerical experiments}\label{S:numerical_experiments}

\begin{figure*}
	\centering
	\begin{minipage}[t]{.34\textwidth} 
        \begin{tikzpicture}[baseline,scale=1]

\pgfplotstableread{data/data_exp3.csv}\errtable

\pgfmathsetmacro{\opacity}{0.3}
\pgfmathsetmacro{\contourop}{0.25}

\begin{semilogxaxis}[
    xlabel={(a) Number of samples},
    xmin=201,
    xmax=50000,
    ylabel={Normalized error},
    ymin = .19,
    ymax = .8,
    grid style=densely dashed,
    grid=both,
    legend style={
        at={(1, 1)},
        anchor=north east},
    legend columns=2,
    width=180,
    height=145,
    ]

    \addplot[white!50!red, densely dotted] table [x=samples, y={GSR-O-1}] {\errtable};
    \addplot[white!50!blue, densely dotted] table [x=samples, y={GSR-OH-1}] {\errtable};
    
    \addplot[white!30!red, densely dashed] table [x=samples, y={GSR-O-10}] {\errtable};
    \addplot[white!30!blue, densely dashed] table [x=samples, y={GSR-OH-10}] {\errtable};
    
    \addplot[black!10!red, solid] table [x=samples, y={GSR-O-100}] {\errtable};
    \addplot[black!10!blue, solid] table [x=samples, y={GSR-OH-100}] {\errtable};
    
    \addplot[white] table [x=samples, y={Offline}] {\errtable};
    
    \addplot[white!10!green, solid] table [x=samples, y={Offline}] {\errtable};
    
    \legend{{OnST-1}, {OnST-H-1}, {OnST-10},{OnST-H-10}, {OnST-100},{OnST-H-100},{ }, {OffST-H}}
\end{semilogxaxis}
\end{tikzpicture}
	\end{minipage}%
	\begin{minipage}[t]{.34\textwidth} 
        \begin{tikzpicture}[baseline,scale=1]

\pgfplotstableread{data/data_exp2.csv}\errtable

\pgfmathsetmacro{\opacity}{0.3}
\pgfmathsetmacro{\contourop}{0.25}

\begin{semilogxaxis}[
    xlabel={(b) Number of samples},
    xmin=100,
    xmax=10000,
    ylabel={Normalized error},
    ymin = .17,
    ymax = .63,
    grid style=densely dashed,
    grid=both,
    legend style={
        at={(1, 1)},
        anchor=north east},
    legend columns=1,
    width=180,
    height=145,
    ]

    \addplot[black!10!red, solid] table [x=samples, y={GSR-O H=2}] {\errtable};
    \addplot[white!30!red, densely dotted] table [x=samples, y={GSR-O H=5}] {\errtable};
    
    \addplot[black!10!blue, solid] table [x=samples, y={GSR-OH H=2}] {\errtable};
    \addplot[white!30!blue, densely dotted] table [x=samples, y={CSR-OH H=5}] {\errtable};
    
    \addplot[white!10!green, solid] table [x=samples, y={Offline H=2}] {\errtable};
    \addplot[white!30!green, densely dotted] table [x=samples, y={Offline H=5}] {\errtable};
    
    \legend{{OnST, $H=2$}, {OnST, $H=5$},{OnST-H, $H=2$},{OnST-H, $H=5$},{OffST, $H=2$},{OffST, $H=5$}}
\end{semilogxaxis}
\end{tikzpicture}	
	\end{minipage}%
	\begin{minipage}[t]{.34\textwidth} 
        \begin{tikzpicture}[baseline,scale=1]

\pgfplotstableread{data/data_exp_real.csv}\errtable

\pgfmathsetmacro{\opacity}{0.3}
\pgfmathsetmacro{\contourop}{0.25}

\begin{axis}[
    xlabel={},
    xtick={},
    ylabel={Stock values},
    ylabel style={rotate=180},
    ylabel shift={-12},
    axis y line*=right,
    ytick pos=right,
    xtick pos=left,
    xmin=20,
    xmax=1205,
    ymin = -1.5,
    ymax = 1.5,
    %
    legend style={
        at={(1, 1)},
        anchor=north east},
    legend columns=2,
    width=180,
    height=145,
    ]

    \addplot[white!70!black, solid] table [x=samples, y={fin_data}] {\errtable};
\end{axis}

\begin{axis}[
    xlabel={(c) Number of samples},
    ylabel={Normalized error},
    xmin=20,
    xmax=1205,
    ymin = -0.1,
    ymax = 0.5,
    axis y line*=left,
    ytick pos=left,
    xtick pos=left,
    grid style=densely dashed,
    grid=both,
    legend style={
        at={(1, 1)},
        anchor=north east},
    legend columns=3,
    width=180,
    height=145,
    ]
    
    \addplot[red, solid] table [x=samples, y={OnST}] {\errtable};
    \addplot[blue, solid] table [x=samples, y={OnST-H}] {\errtable};

    \legend{OnST, OnST-H, {Stock values}}
   \addlegendimage{white!70!black, solid}
\end{axis}

\end{tikzpicture}	
	\end{minipage}
 
    \vspace{-0.2cm}
	\caption{Normalized graph estimation error ${\rm err}(\bbSo)$ versus the number of samples while considering 3 approaches combined with (a) 3 values for the number of iterations $\{1,10,100\}$ performed at each time $t$ for OnST, OnST-H, (b) 2 values for the hidden nodes $H = \{2,5\}$. (c) ${\rm err}(\bbSo)$ for OnST and OnST-H (left $y$-axis) and standardized average stock values (right $y$-axis) versus the number of samples.}
	\label{F:exp_123}
\end{figure*}
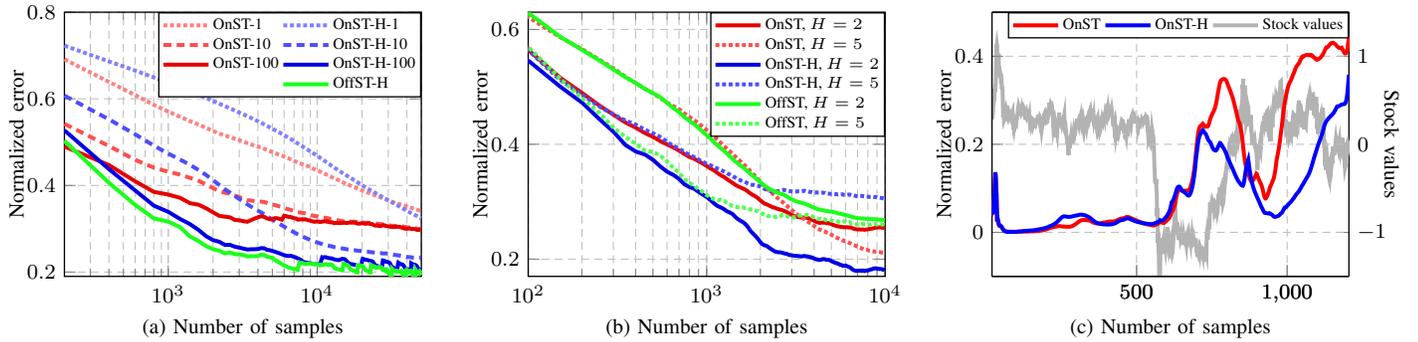

In this section, we conduct several synthetic and real data experiments to evaluate the performance of the proposed approach in scenarios involving online settings and the presence of hidden nodes.
The different approaches considered for the experiments are:
\begin{itemize}[left= 5pt .. 15pt, noitemsep]
    \item An online graph learning approach from streaming stationary graph signals~\cite{alglearninggraphs}. We refer to this method as ``OnST''.
    \item 
    Our proposed approach, denoted ``OnST-H'', using Alg.~\ref{alg:PG} for online graph learning while accounting for hidden nodes. 
    \item An offline approach to learn graphs from streaming data with hidden nodes, which optimizes~\eqref{e:opt_online} for each new signal at time $t$ by performing a large number of proximal gradient iterations. We refer to this method as ``OffST-H''. 
\end{itemize}
To measure the quality of the estimated graphs, we use the normalized squared Frobenius norm defined as 
\begin{alignat}{2}\label{nfronorm}
    {\rm err}(\hbSo) = \| \bbSo^*-\hbSo \|_F^2/\|\bbSo^*\|_F^2,
\end{alignat}
where $\bbSo^*$ and $\hbSo$ stand for the target and estimated $\bbSo$, respectively.
Further implementation details of our ensuing simulations, as well as results for estimation metrics other than \eqref{nfronorm}, are available in the online code repository {\url{https://github.com/andreibuciulea/OnlineStHidd}}.

\medskip

\noindent \textbf{Online graph estimation.} 
We first evaluate the performance of our approach ``OnST-H'' compared to the baseline methods ``OnST'' and ``OffST-H'' for learning graphs from streaming data in the presence of hidden nodes.
We generate stationary graph signals using a covariance matrix as a polynomial of the GSO $\bbC = (\sum_{l=0}^{L-1} h_l \bbS^l)^2$ with coefficients $h_l$ sampled from a Gaussian distribution.
The underlying graph structure is modeled as an Erdős-Rényi (ER) graph with $N = 30$ nodes and an edge probability of $p = 0.1$, with $H = 2$ hidden nodes selected uniformly at random.
Fig. \ref{F:exp_123}a reports ${\rm err}(\hbSot)$ on the $y$-axis and the number of samples $t$ on the $x$-axis. 
The number following each method name in the legend indicates the number of times that steps 4 to 9 of Alg.~\ref{alg:PG} are repeated per time step $t$.

We observe that increasing the number of proximal gradient iterations per sample improves the performance of both OnST and OnST-H. 
Indeed, performing only one iteration per time step results in higher error. 
However, when we increase the number of iterations to 10 or 100 per sample, the estimation error approaches that of the offline method, albeit at the cost of additional computation time. 
This highlights the necessity of seeking a trade-off between computation time and estimation accuracy, where 10 iterations per sample yields a suitable balance between performance and efficiency.
Regarding the impact of hidden nodes, we also observe that with a sufficiently large number of samples ($\geq 3 \cdot 10^3$), OnST-H outperforms OnST, particularly when the number of iterations per sample is 10 or more.

\medskip

\noindent \textbf{Influence of hidden nodes.}
The goal of this experiment is to quantify the impact of the number of hidden variables in online graph learning settings. 
To achieve this, we generate stationary signals and assess the performance of each method for different values of $H$ as the number of samples grows. 
Specifically, we employ Erdős–Rényi (ER) graphs with $N = 20$ nodes and an edge probability of $p = 0.1$. 
The results in Fig. \ref{F:exp_123}b report the average estimation error for each method, with $H=2$ and $H=5$ hidden nodes represented by solid and dotted lines, respectively, for an increasing number of samples. 

Focusing on the effect of the number of hidden nodes, we see that both OnST and OnST-H achieve lower estimation errors when  $H=2$ compared to $H=5$. This observation aligns with theoretical expectations, as a higher number of hidden nodes increases the complexity of the problem, leading to larger estimation errors. Additionally, examining the importance of accounting for the hidden variables, we observe that for a sufficiently large number of samples, OnST-H consistently outperforms OnST in both scenarios with $H=2$ and $H=5$, highlighting the advantages of explicitly accounting for hidden variables in the estimation process.
Finally, note that the error of OnST-H is much closer than that of OnST to the ideal offline method OffST-H for $H=2$ and $H=5$.

\medskip

\noindent \textbf{Financial data.}
Finally, we test the performance of our proposed approach on a real-world financial dataset~\cite{money2023scalable}.
We are given financial signals associated with $N=15$ companies from the SP\&500 from June 1st, 2019, to October 14th, 2022 with a total of $M=1206$ signals, and we consider $H=2$ hidden nodes.
To serve as our ground truth time-varying graphs, we estimate a graph at each time $t$ using all $N=15$ nodes in an offline manner from all $t$ available signals, and we compare the sub-graphs on the observed nodes to our online estimates.
Fig.~\ref{F:exp_123}c shows the normalized estimation error on the left $y$-axis comparing OnST and OnST-H for an increasing number of signals, along with the standardized average of stock values on the right $y$-axis.
We see that OnST-H is better able to track the change in the graph topology encoding financial similarities between companies, even when the data experiences sharp changes in distribution due to the COVID-19 pandemic.
Thus, we observe that hidden information in real-world streaming data can have detrimental effects on time-varying graph estimation, but our method is able to mitigate these errors even under significant changes in data distribution.


\section{Conclusions}\label{S:conclusions}
We proposed a method to learn graphs from stationary graph signals in an online scenario where some nodes are never observed.
We formulated an optimization problem to estimate a dynamic graph from streaming signals, and we modeled the influence of hidden nodes under the assumption of stationarity.
We proposed an efficient proximal gradient algorithm with guarantees that online estimation can compete with batch-wise estimation.
Moreover, the algorithm can also be applied in an offline setting, providing a more efficient approach than existing works that consider hidden nodes.
Our results show that accounting for hidden nodes in online scenarios is crucial for estimating a meaningful graph and for downstream tasks.

\vspace{1cm}

\vfill\pagebreak
\bibliographystyle{IEEE}
\bibliography{citations}

\end{document}